\documentclass[conference]{IEEEtran}
\usepackage{cite}
\usepackage{amsmath,amssymb,amsfonts}
\usepackage{algorithmic}
\usepackage[dvipdfmx]{graphicx}
\usepackage{textcomp}
\usepackage{xcolor}
\usepackage{tikz}

\usepackage{multirow} 

\newcommand\copyrighttext{%
  \footnotesize \textcopyright 2024 IEEE. Personal use of this material is permitted. Permission from IEEE must be obtained for all other uses, including reprinting/republishing this material for advertising or promotional purposes, collecting new collected works for resale or redistribution to servers or lists, or reuse of any copyrighted component of this work in other works.}

\newcommand\copyrightnotice{%
\begin{tikzpicture}[remember picture,overlay]
\node[anchor=south,yshift=10pt] at (current page.south) {\fbox{\parbox{\dimexpr0.75\textwidth-\fboxsep-\fboxrule\relax}{\copyrighttext}}};
\end{tikzpicture}%
}

\renewcommand\fbox{\fcolorbox{red}{white}}
\def\BibTeX{{\rm B\kern-.05em{\sc i\kern-.025em b}\kern-.08em
    T\kern-.1667em\lower.7ex\hbox{E}\kern-.125emX}}
\begin{document}

\title{Effects of the retina-inspired light intensity encoding on color discrimination performance\\
}


\author{\IEEEauthorblockN{1\textsuperscript{st} Io YAMADA}
\IEEEauthorblockA{\textit{Graduate School of Information Science and Technology} \\
\textit{Osaka Institute of Technology}\\
Osaka, Japan \\
m1m23a37@st.oit.ac.jp}
\and
\IEEEauthorblockN{2\textsuperscript{nd} Hirotsugu OKUNO}
\IEEEauthorblockA{\textit{Faculty of Information Science and Technology} \\
\textit{Osaka Institute of Technology}\\
Osaka, Japan \\
hirotsugu.okuno@oit.ac.jp}
}


\maketitle
\copyrightnotice
\begin{abstract}
Color is an important source of information for visual functions such as object recognition, but it is greatly affected by the color of illumination. The ability to perceive the color of a visual target independent of illumination color is called color constancy (CC), and is an important feature for vision systems that use color information. In this study, we investigated the effects of the light intensity encoding function on the performance of CC of the center/surround (C/S) retinex model, which is a well-known model inspired by CC of the visual nervous system. The functions used to encode light intensity are the logarithmic function used in the original C/S retinex model and the Naka-Rushton (N-R) function, which is a model of retinal photoreceptor response. Color-variable LEDs were used to illuminate visual targets with various lighting colors, and color information computed by each model was used to evaluate the degree to which the color of visual targets illuminated with different lighting colors could be discriminated. Color information was represented using the HSV color space and a color plane based on the classical opponent color theory. The results showed that the combination of the N-R function and the double opponent color plane representation provided superior discrimination performance.
\end{abstract}


\section{Introduction}
\label{sec:Introduction}
Color information is used in a wide range of applications in computer and robot vision, such as object recognition. However, since the light intensity received by an image sensor is expressed as the product of illumination intensity and object reflectance, the color information can easily change depending on the illumination intensity and color. In order to stably acquire the true color information of an object, a mechanism to reduce the influence of illumination light is necessary.

The ability to perceive the color of a visual target independent of the illumination color is called color constancy (CC), and a number of computational CC algorithms have been developed in the field of computer vision (see \cite{gijsenij_computational_2011} for an overview). Many computational CC algorithms consist of two steps: estimating the lighting color from images obtained under unknown lighting conditions and generating an image that would be obtained under white lighting. These CC algorithms range from those that rely on relatively simple statistics (e.g., \cite{buchsbaum_spatial_1980, land_retinex_1977, van_de_weijer_edge-based_2007}) to those that are learning-based (e.g., \cite{forsyth_novel_1990, brainard_bayesian_1997, gijsenij_color_2007}). While these methods have been evaluated based on accuracy in estimating illumination color using image datasets \cite{gijsenij_computational_2011}, applications for robot vision, such as target tracking and object recognition, require measurement of the performance of target color discrimination under changing lighting conditions.

On the other hand, the visual nervous system (VNS) of humans and animals can perceive the color of the object itself with minimal influence of illumination. The retinex model \cite{land_lightness_1971} is widely known as a computational model of the VNS that achieves CC. The retinex model is relatively computationally inexpensive and, unlike the computational CC algorithms described above, can estimate the color of the visual target without performing illumination color estimation. For this reason, this model is suitable to robot vision.

The center/surround (C/S) retinex model, which combines the retinex model with a C/S antagonistic spatial filter used in the VNS, has been proposed as a realistic model for achieving CC \cite{jobson_properties_1997, land_alternative_1986}. This model showed that applying logarithmic compression and a C/S antagonistic filter such as a difference of Gaussians (DoG) filter to the red (R), green (G), and blue (B) channels of the input image is effective in achieving CC. A model that extends the Gaussian filter of this model to multi-scale has also been proposed \cite{jobson_multiscale_1997}. A retinex model in which the retinal mechanisms were more closely simulated has also been developed and has shown good illumination color estimation accuracy \cite{zhang_retinal_2016}.

Some hardware implementations of a part of the C/S retinex model have been made \cite{moore_real-time_1991, shimonomura_wide-dynamic-range_2011}. An image sensor system with CC was also developed using an FPGA with the C/S retinex model being implemented \cite{misaka_fpga_2020}. These efforts are the first step toward developing retinex algorithms for robot vision applications that require real-time processing and compactness. On the other hand, the application to robot vision requires careful consideration of how to transform light intensity, which varies widely in real world environments, into an input signal.

The relationship between input light intensity and the response potential of photoreceptor cells, which are the input cells of the biological VNS, is often approximated by a logarithmic function, and the C/S retinex model also uses logarithmic transformation to encode the light intensity. On the other hand, the Naka-Rushton (N-R) function is often used as a more accurate model of photoreceptor characteristics \cite{naka_s-potentials_1966}. From an engineering point of view, the logarithmic transformation of an image assigns more gray levels to low-luminance information, leading to the problem of insufficient levels for information above a certain luminance depending on the lighting environment. Therefore, the encoding method of light intensity could have an influence on the color discrimination performance.

Although from a different perspective than CC performance, the C/S retinex model was used to perform tone mapping of high dynamic range (HDR) images, and the relationship between the functions used in the model and the quality of the output images was investigated \cite{lisani_analyzing_2020}. The N-R function was used as one of the functions that convert an HDR image into an 8-bit image, and the results showed that the N-R function was effective in generating images with high visual quality.

A CC algorithm in which the logarithmic transformation of the C/S retinex model is replaced by the N-R function (hereinafter referred to as the N-R C/S retinex model) has also been proposed \cite{hisamitsu_image_2022}. Changes in hue and saturation of visual targets computed by the proposed model during two different outdoor time periods were investigated. The results suggested that the use of the N-R function may reduce the amount of hue change. However, the color discrimination performance of the N-R C/S retinex has not been investigated.

Color discrimination performance inevitably depends on the color representation method. The most widely used color space that can be easily converted from RGB data is the HSV color space, where colors are represented by hue, saturation, and value (brightness). On the other hand, a color plane based on the classical opponent color theory has long been used, which is an orthogonal color plane with red-green (RG) and blue-yellow (BY) axes \cite{shevell_color_2017}, hereinafter referred to as the double opponent (DO) color plane. In fact, it is well-known that color information in the VNS is transmitted by color opponent cells, which respond to RG and BY opponency \cite{shapley_color_2011}. A combination of the neuro-inspired encoding method, CC algorithm, and color plane could improve color discrimination accuracy.

The purpose of this study is to evaluate the color discrimination performance of the C/S retinex model in terms of robot vision applicability. We investigated the performance of the model in discriminating one color target from another color in the presence of large changes in lighting conditions that could occur in a real environment. The models used are the C/S retinex model with the conventional logarithmic function, the C/S retinex model with the N-R function, and the gray world (GW) algorithm \cite{buchsbaum_spatial_1980}. The color spaces used are the HSV and the DO color plane. To evaluate the discrimination performance, we used the Fisher criterion (FC) \cite{noauthor_pattern_nodate}.
\section{Algorithms for color constancy}
\label{sec:Algorithms for color constancy}
\subsection{Center/surround retinex model}
\label{sbsec:Center/surround retinex model}
\begin{figure}[b]
    \centering
		\includegraphics[keepaspectratio=true,width=0.8\linewidth]{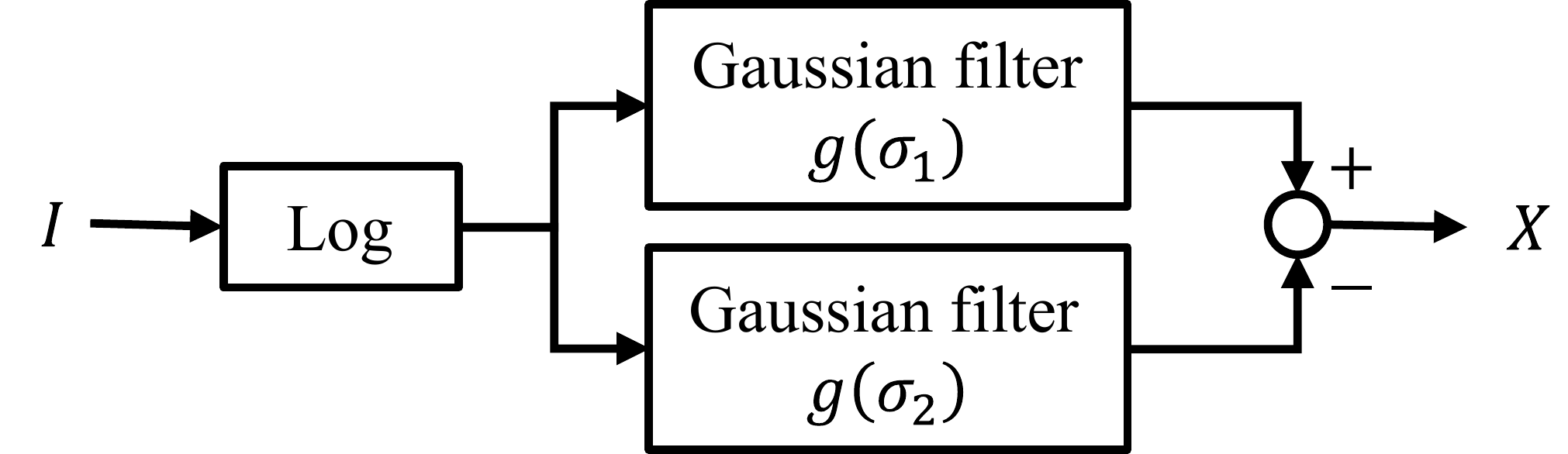}
	\caption{Processing flow diagram of the center/surround retinex model for a single channel.}
	\label{fig:flow_CSRetinex}
\end{figure}

Fig. \ref{fig:flow_CSRetinex} shows the processing flow of the C/S retinex model. The algorithm of this computational model consists of the following three processes: logarithmic transformation, Gaussian filtering, and subtraction of two Gaussian filtered images. This process is performed separately on each of the three-color (RGB) channels.

First, the light intensity $I$ acquired by an image sensor is compressed logarithmically. Here we consider light coming from an ideal diffusely reflecting surface; the light is expressed as the product of the illumination intensity $E$ and the surface reflectance $\rho$. The output of the logarithmic compression is expressed as
\begin{equation}
\begin{split}
L(\boldsymbol{x}) &= \log I(\boldsymbol{x})\\
&= \log E(\boldsymbol{x}) + \log \rho(\boldsymbol{x}),
\end{split}
\label{equ:log_simple}
\end{equation}
where $\boldsymbol{x} = (x,y)$ represents the coordinates of the image. According to equation (\ref{equ:log_simple}), the input light intensity $I$ is expressed in the sum of terms with $E$ and $\rho$.

Next, Gaussian filters $g(\boldsymbol{x};\sigma)$ with two different standard deviations (SD) $(\sigma_1, \sigma_2 (\sigma_1 < \sigma_2))$ are applied to the output of logarithmic compression. Finally, the difference between the two Gaussian filtered images is computed by
\begin{equation}
\begin{split}	
X(\boldsymbol{x}) &= g(\boldsymbol{x}; \sigma_{1}) * L(\boldsymbol{x}) - g(\boldsymbol{x}; \sigma_{2}) * L(\boldsymbol{x})\\
&= g(\boldsymbol{x}; \sigma_{1}) * \log E(\boldsymbol{x}) + g(\boldsymbol{x}; \sigma_{1}) * \log \rho(\boldsymbol{x})\\
&- g(\boldsymbol{x}; \sigma_{2}) * \log E(\boldsymbol{x}) - g(\boldsymbol{x}; \sigma_{2}) * \log \rho(\boldsymbol{x}).
\end{split}
\label{equ:C/S Retinex}
\end{equation}
Assuming that the illumination intensity $E$ is uniformly spread over the range of the Gaussian filter, terms involving $E$ cancel each other out by subtraction. In this way, the effect of illumination light is reduced by the C/S retinex model.

The set of SD used in this study was ($\sigma_1$, $\sigma_2$) = (1.057, 17.964) pixels. Spatial filters with large kernels are inherently computationally expensive and therefore unsuitable for robot vision, but Gaussian filters can be applied efficiently using hierarchical discrete correlation \cite{burt_fast_1981, yamaji_fast_2022}.

\subsection{Logarithmic transformation}
\label{sbsec:Logarithmic transformation}
Equation (\ref{equ:log_detail}) expresses the logarithmic transformation in the C/S retinex model used in this study.
\begin{align}
\label{equ:log_detail}
L(\boldsymbol{x}) &= \beta (\log_{2} (I(\boldsymbol{x}) - \alpha) - \gamma)
\\
\label{equ:log_alpha}
\alpha &= I_{\min} - 2^{\gamma}
\\
\label{equ:log_beta}
\beta &= \dfrac{255}{\log_{2} (I_{\max} - \alpha) - \gamma}
\end{align}
Parameters $\alpha$ and $\beta$ are set so that the converted values are distributed from 0 to 255 because the image data are represented by 8-bit values. Parameter $\gamma$ is used to change the compression ratio of the logarithmic transformation.
$I_{\min}$ and $I_{\max}$ represent the minimum and maximum pixel values to be converted.
Values for $I_{\min}$ and $I_{\max}$ were determined by the cumulative histogram $H_{c}(k)$ of the input image, where $k$
is the pixel value, instead of using the actual minimum and maximum pixel values for $I_{\min}$ and $I_{\max}$ directly because these values can fluctuate greatly for each frame.
For convenience, the minimum values of $k$ that satisfy inequality $H_{c}(k) > N_{T} / 256$ and $H_{c}(k) > N_{T} - N_{T} / 256$ were used for $I_{\min}$ and $I_{\max}$, respectively.
Here, $N_{T}$ represents the total number of pixels. In this study, $N_{T}$ = 19200.

\begin{figure}[bt]
    \centering
		\includegraphics[keepaspectratio=true,width=0.75\linewidth]{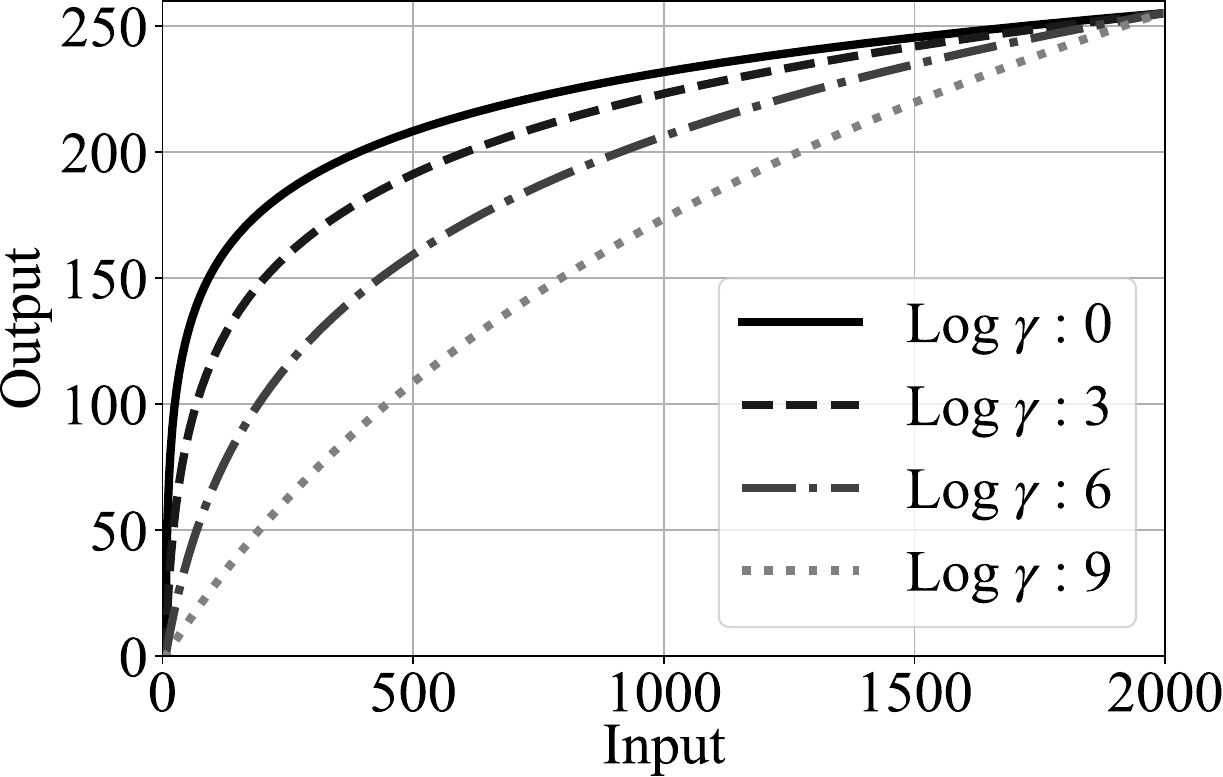}
	\caption{Input-output relationship of logarithmic transformation for various values of parameter $\gamma$.}
	\label{fig:comp_Log_gamma}
\end{figure}

Fig. \ref{fig:comp_Log_gamma} shows the input-output relationship of the logarithmic transformation for various values of parameter $\gamma$. When the parameter $\gamma$ is small, more tones are assigned to the representation of low-luminance tones of the input. This results in a lack of available tones for color representation when the target object is in the medium or high intensity range. In contrast, when the parameter $\gamma$ is large, the curve approaches linearity and the logarithmic property is lost.

\subsection{Naka-Rushton center/surround retinex model}
\label{sbsec:Naka-Rushton center/surround retinex model}
\begin{figure}[b]
    \centering
		\includegraphics[keepaspectratio=true,width=0.8\linewidth]{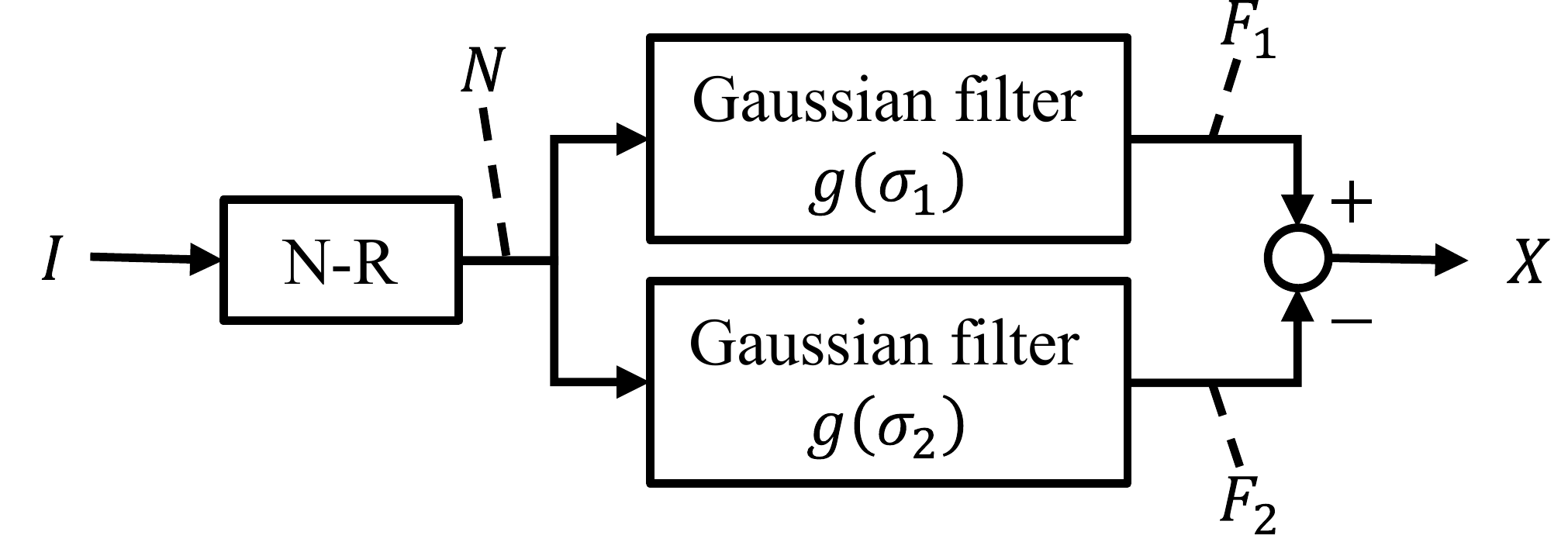}
	\caption{Processing flow diagram of the center/surround retinex model that uses Naka-Rushton function for a single channel.}
	\label{fig:flow_NRRetinex}
\end{figure}

In this study, we investigated the output of a modified C/S retinex model that uses the N-R function instead of the logarithmic function. Fig. \ref{fig:flow_NRRetinex} shows the processing flow of the N-R C/S retinex model.

The N-R function expresses the relationship between the light intensity received by photoreceptor cells in the biological retina and its neural response, and the response $N$ is expressed as
\begin{equation}
\label{equ:NR_withN}
N = V_{m}\dfrac{I^{n}}{I^{n} + I^{n}_{h}},
\end{equation}
where $V_{m}$ is the maximum response, $I$ is the input light intensity, and $I_{h}$ is the light intensity at which the neural response reaches half of the maximum response. Parameter $n$ is a fitting parameter that is approximately from 0.7 to 1 for most animals and humans. In this study, parameter $n$ was set to 1.
The N-R function with $n = 1$ can be approximated as a logarithmic function around $I = I_{h}$, which allows for the reduction of illumination effects over a certain range of $I$ around $I_{h}$.
$I_{h}$ was set to the median value of all pixels in each color channel.

\begin{figure}[bt]
    \centering
		\includegraphics[keepaspectratio=true,width=0.75\linewidth]{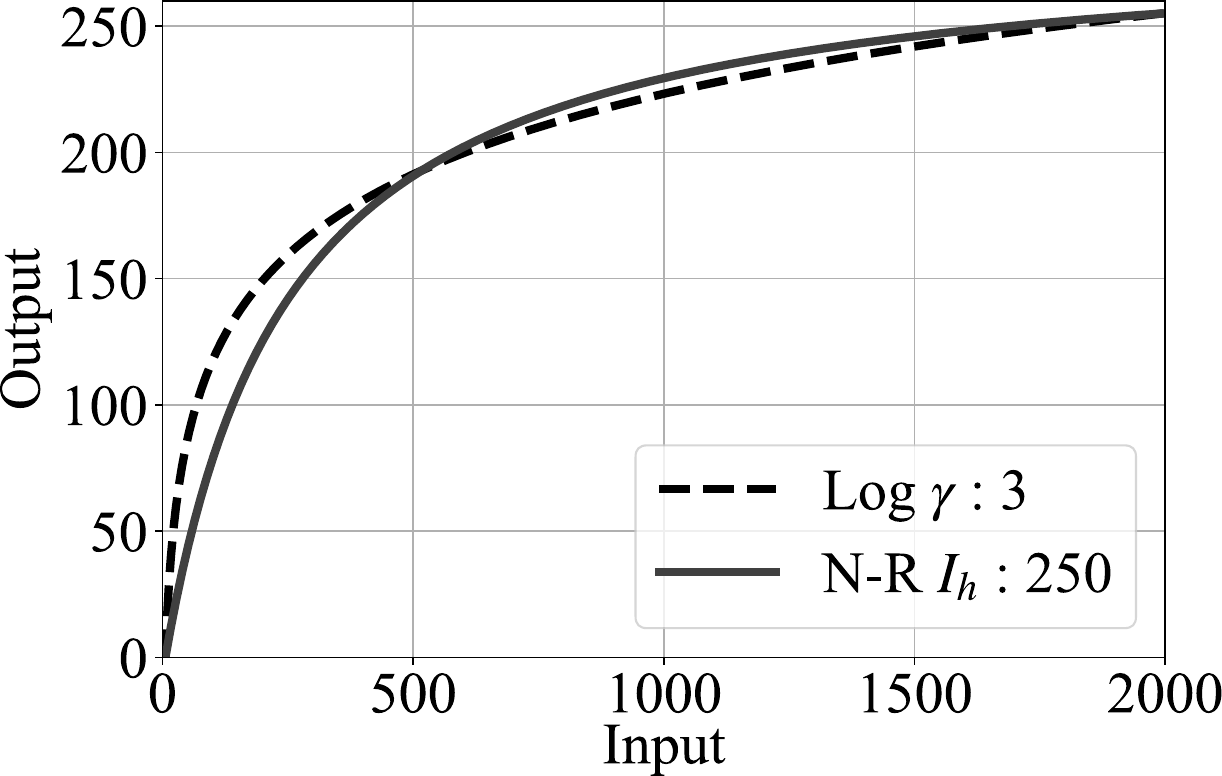}
	\caption{Input-output relationship for the logarithmic function and the N-R function.}
	\label{fig:comp_Log_N-R}
\end{figure}

Fig. \ref{fig:comp_Log_N-R} shows the input-output relationship for the logarithmic function and the N-R function. Fig. \ref{fig:comp_Log_N-R} shows that the N-R function expresses fewer tones in the low-luminance range and more tones in the medium to high intensity range than the logarithmic function.
This property improves the problem of the lack of tones representing objects in a higher intensity area.

\begin{figure}[bt]
    \centering
		\includegraphics[keepaspectratio=true,width=0.85\linewidth]{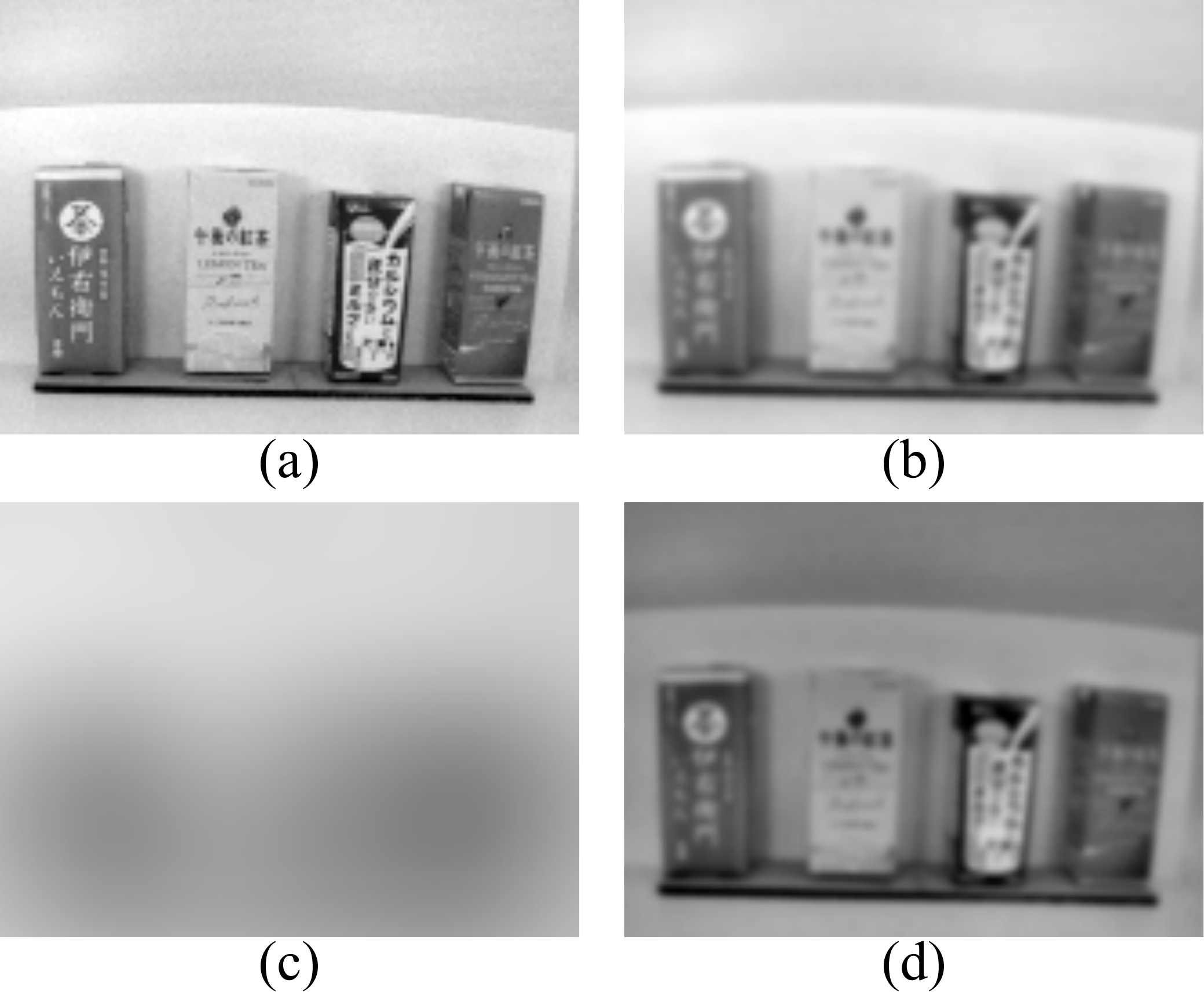}
	\caption{Images in the process of computation of the N-R C/S retinex model in the R channel. (a) Output image of the N-R function indicated by $N$ in Fig. \ref{fig:flow_NRRetinex}. (b)(c) Gaussian filtered images indicated by $F_{1}$ and $F_{2}$ in Fig. \ref{fig:flow_NRRetinex}, respectively. (d) Output image of this channel indicated by $X$ in Fig. \ref{fig:flow_NRRetinex}.}
	\label{fig:NR_intermediateImage}
\end{figure}

Equation (\ref{equ:NR_ImplementedVer}) is the implemented form of the function.
\begin{equation}
\label{equ:NR_ImplementedVer}
N = 
\begin{cases}
V_{m}\dfrac{I - I_{\min}}{(I - I_{\min}) + I_{h}},		& \text{for $I_{\min} \leq I \leq I_{\max}$} \\
0,														& \text{for $I < I_{\min}$} \\
255,														& \text{for $I > I_{\max}$}
\end{cases}
\end{equation}
Here, $I_{\min}$ and $I_{\max}$ represent the minimum and maximum pixel values to be converted. These values are the same as the previous section. $V_{m}$ is determined so that the maximum value of $N$ becomes 255 and is given by:
\begin{equation}
\label{equ:NR_Vm}
V_{m} = \dfrac{255(I_{\max} - I_{\min} + I_{h})}{I_{\max} - I_{\min}}.
\end{equation}
Example images in process of computation of the N-R C/S retinex model are shown in Fig. \ref{fig:NR_intermediateImage}.

\subsection{Gray world algorithm}
\label{sbsec:gray world algorithm}
The GW algorithm adjusts white balance based on the assumption that the average reflectance of surfaces in the world is achromatic, and is widely used as a low-computational-cost method for color correction. Therefore, we used the GW algorithm as a performance comparison for CC. Equation (\ref{equ:GW_equG}) expresses the output of the GW algorithm for a green channel $X_G$ used in this study.
\begin{equation}
\label{equ:GW_equG}
X_{G} = \frac{128}{\mu_{G}} I_{G}
\end{equation}
Here, $I_G$ denotes the input, and the variable $\mu_G$ denotes the average value of the green channel. The same process is applied to the red and blue channels. This calculation unifies the mean values for each channel to 128.
\section{Evaluation methods}
\label{sec:Evaluation methods}
\subsection{Color space}
\label{sbsec:Color space}
The color discrimination performance was evaluated using values represented in the HSV or DO color space. The output RGB signals obtained by the CC algorithms described above were converted into these color spaces. 

The hue ($H$), saturation ($S$), and brightness ($V$) in the HSV color space are obtained by the following equations.
\begin{align}
\label{equ:RGBtoHSV_H}
H &= 
	\begin{cases}
		\dfrac{60(G-B)}{V - \min(R,G,B)},		& \text{for $V = R$} \\
		120 + \dfrac{60(B-R)}{V - \min(R,G,B)},	& \text{for $V = G$} \\
		240 + \dfrac{60(R-G)}{V - \min(R,G,B)},	& \text{for $V = B$} \\
		0,										& \text{for $R = G = B$}
	\end{cases}
 \\
 \label{equ:RGBtoHSV_S}
S &= 
	\begin{cases}
		\dfrac{V - \min(R,G,B)}{V},	& \text{for $V \neq 0$} \\
		0,							& \text{for $V = 0$}
	\end{cases}
\\
\label{equ:RGBtoHSV_V}
V &= \max(R,G,B)
\end{align}
The variables $R$, $G$, and $B$ represent the pixel values of each RGB channel. The ranges of hue ($H$), saturation ($S$), and brightness ($V$) are $0 \leq H < 360$, $0 \leq S \leq 255$, and $0 \leq V \leq 255$, respectively. If $H$ is less than 0, 360 is added to $H$.

In addition to the HSV color space, we used the DO color plane based on the classical opponent color theory; the color plane is an orthogonal color plane with RG and BY axes. The values along RG and BY axes are calculated by the following equation:
\begin{align}
\label{equ:RGBtoOPPO_RG}
O_{RG} &= X_{R} - X_{G}
\\
\label{equ:RGBtoOPPO_YB}
O_{YB} &= \dfrac{X_{R} + X_{G}}{2} - X_{B}
\end{align}
In accordance with the polarity of the original DO color plane, we set yellow to be positive and blue to be negative on the BY opponent color axis. The values represented in polar coordinates in this color plane correspond to hue and saturation.
\begin{align}
\label{equ:OPPOtoOPPOpc_r}
r &= \sqrt{O^{2}_{RG} + O^{2}_{YB}}
\\
\label{equ:OPPOtoOPPOpc_theta}
\theta &= \arctan\dfrac{O_{YB}}{O_{RG}}
\end{align}
$\theta$ corresponds to hue, and $r$ corresponds to saturation.
If $\theta$ is less than 0, 360 is added to $\theta$.

\subsection{Fisher criterion}
\label{sbsec:Fisher criterion}
We used the FC to evaluate the color discrimination performance of each method. This criterion is calculated for two different regions to be compared, and a higher value of FC indicates that the color pair is easier to discriminate. The FC ($D$) is expressed as the ratio of the within-class variance ($s_w^2$) shown in equation (\ref{equ:withinClassVariance}) to the between-class variance ($s_b^2$) shown in equation (\ref{equ:betweenClassVariance}).
\begin{align}
\label{equ:withinClassVariance}
s^{2}_{w} &= s^{2}_{1} + s^{2}_{2}
\\
\label{equ:betweenClassVariance}
s^{2}_{b} &= (\mu_{2} - \mu_{1})^{2}
\\
\label{equ:FisherCriterion}
D &= \dfrac{s^{2}_{b}}{s^{2}_{w}}
\end{align}
Here, variables $\mu_1$ and $\mu_2$ are the means of each of the two different visual targets, and variables $s_1^2$ and $s_2^2$ are the variances for each of the two different visual targets.

\section{Experiments}
\label{sec:Experiments}
\subsection{Experimental environment}
\label{sbsec:Experimental environment}
\begin{figure}[bt]
    \centering
		\includegraphics[keepaspectratio=true,width=0.87\linewidth]{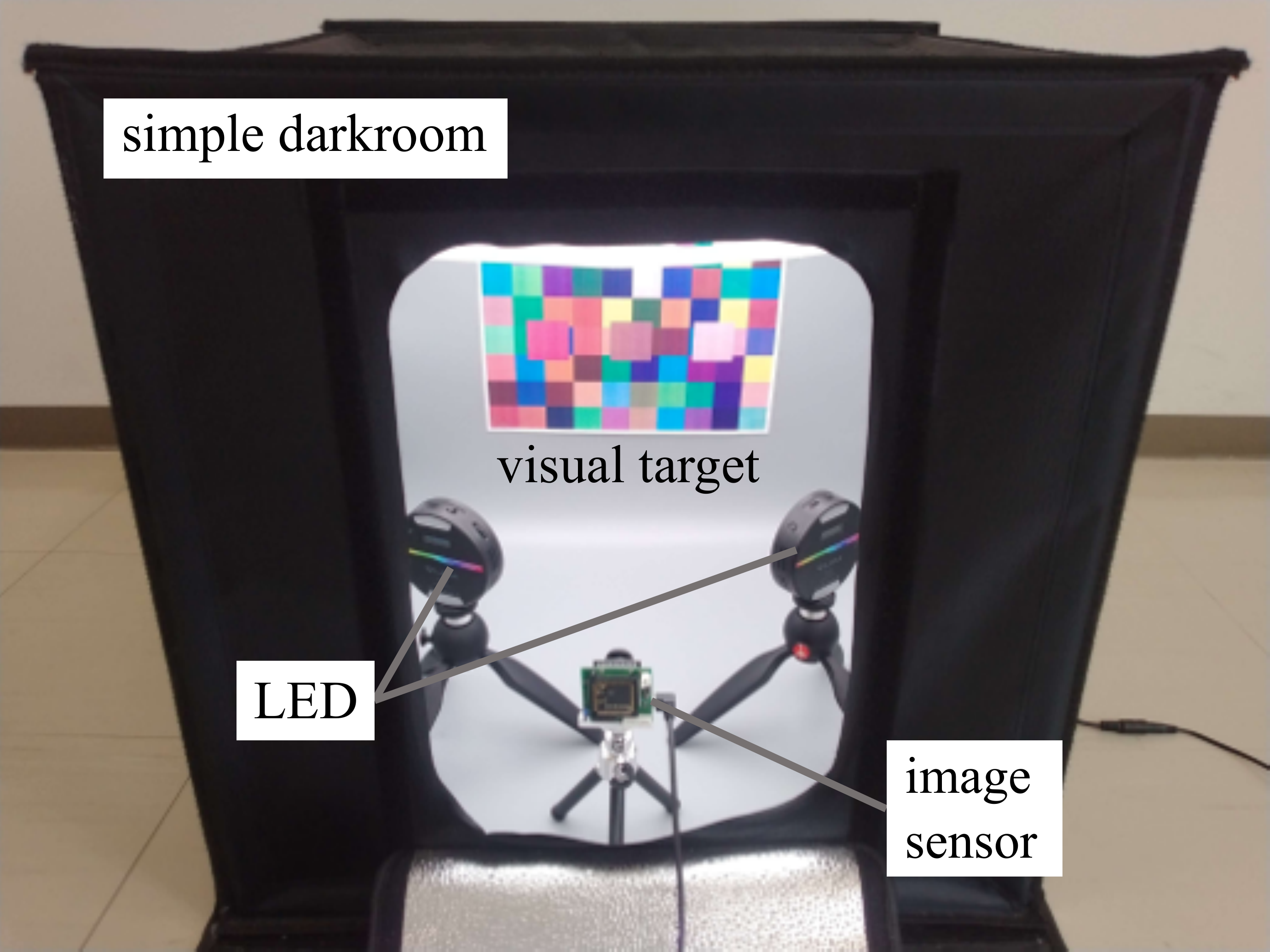}
	\caption{Experimental environment. The image sensor, visual target objects, and a pair of color-variable LEDs for illumination are all placed in a simple darkroom.}
	\label{fig:ExperimentEnvironment}
\end{figure}

We investigated the output of the three methods described in section \ref{sec:Algorithms for color constancy} (C/S retinex model, N-R C/S retinex model, and GW algorithm) using the experimental environment shown in Fig. \ref{fig:ExperimentEnvironment}. An image sensor system, which is explained in section \ref{sbsec:Image sensor system}, visual target objects, which is explained in section \ref{sbsec:Visual target object}, and a pair of color-variable LEDs were all housed in a simple darkroom to minimize the influence of external illumination.

 The light sources used were a white LED installed on the ceiling of the darkroom and two color-variable LEDs located on the left and right sides of the darkroom. During the experiment, the white LED emitted white light of constant intensity, and the two color-variable LEDs emitted four colors (red, yellow, green, and blue) and white light. We used 17 types of colors for illumination as follows: 16 (4 $\times$ 4) color combinations of the left and right variable-color LEDs, and white illumination with all three LEDs white.

\subsection{Image sensor system}
\label{sbsec:Image sensor system}
\begin{figure}[bt]
    \centering
		\includegraphics[keepaspectratio=true,width=1.0\linewidth]{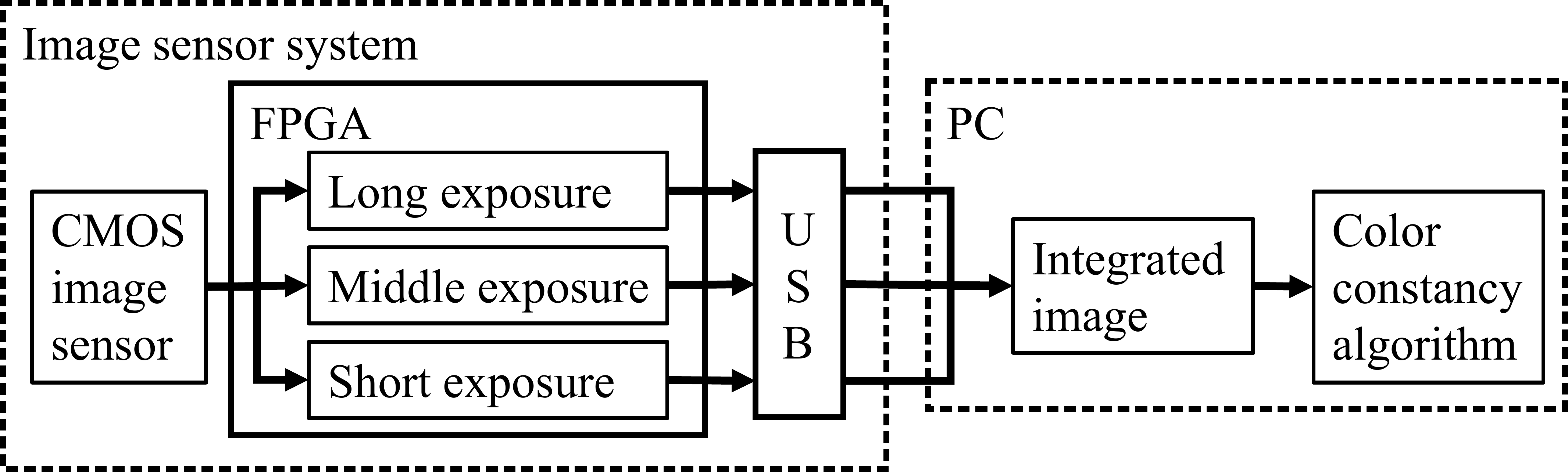}
	\caption{Signal flow diagram of image sensor system.}
	\label{fig:flow_ImageSensor}
\end{figure}

The image sensor system used here is composed of a CMOS image sensor (OmniVision, OV5642) and a field-programmable gate array (FPGA) (AMD, Artix-7 XC7A100T). Fig. \ref{fig:flow_ImageSensor} shows the structure and the signal flow of the system. The image sensor system acquires three 8-bit data images using three different exposure times. The resolution of the images is 160 $\times$ 120 pixels for each color channel. The acquired images were sent to a PC via USB. The images acquired at the three different exposure times were merged into one image by replacing the saturated pixels in the image acquired by a longer exposure time with pixels from the image acquired by a shorter exposure time. Each CC model was applied to this merged image. We used the image whose dynamic range was extended by the multiple exposure technique to avoid the lack of tones in encoding using the logarithmic and N-R functions.

\subsection{Visual target object}
\label{sbsec:Visual target object}
\begin{figure}[bt]
    \centering
		\includegraphics[keepaspectratio=true,width=1.0\linewidth]{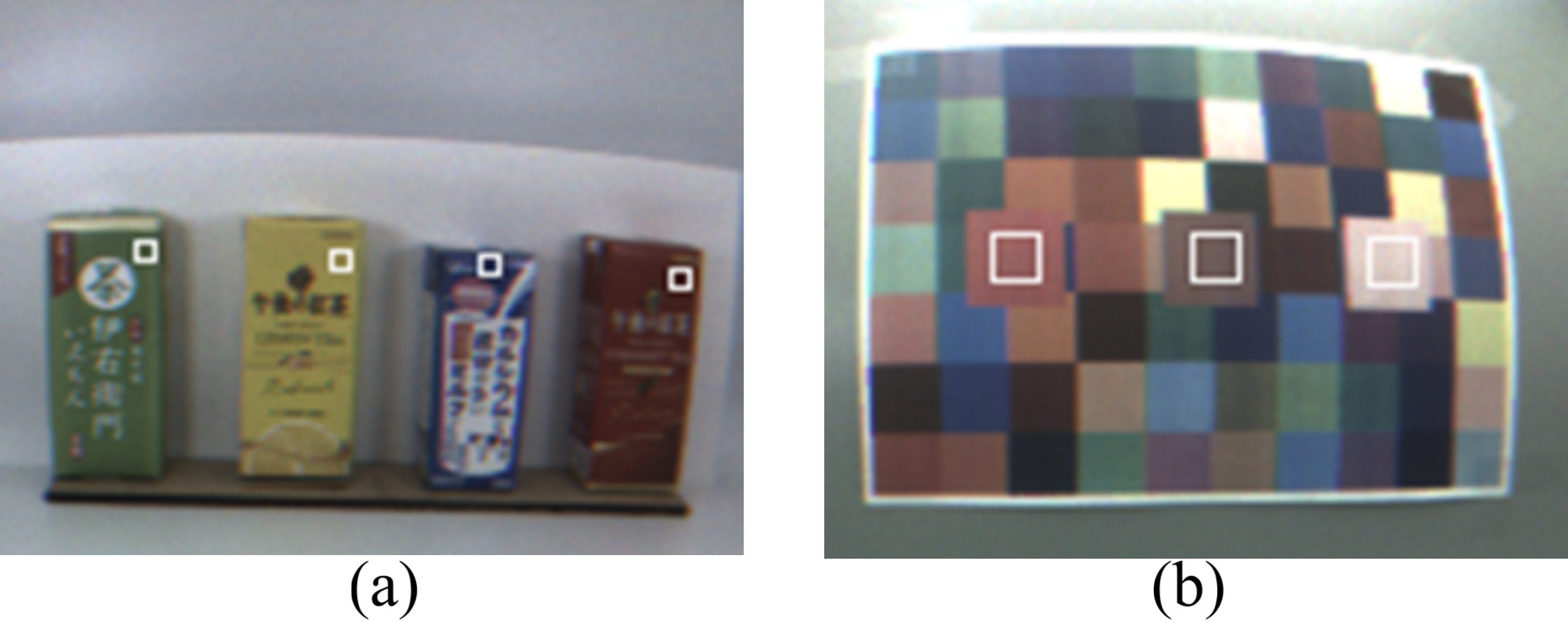}
	\caption{Visual target objects. The colors in the white frames are the target of evaluation. (a) Four-color beverage cartons. (b) Color patches that consisted of randomly colored rectangles printed on a sheet of paper.}
	\label{fig:targetObject}
\end{figure}

The visual target objects used were beverage cartons (Fig. \ref{fig:targetObject}(a)) and color patches that consisted of randomly colored rectangles printed on a sheet of paper (Fig. \ref{fig:targetObject}(b)). The average color of the area surrounded by the white frame in the figure is the target of evaluation.
The size of the white frame in Fig. \ref{fig:targetObject}(a) and (b) are 3 $\times$ 3 pixels, and 10 $\times$ 10 pixels, respectively.

The cartons in four different hues (from left to right: green, yellow, blue, and red) were used to evaluate the ability to discriminate hues.

The color patches were used to evaluate the ability to discriminate saturation and brightness.
The three areas surrounded by the white frames in Fig. \ref{fig:targetObject}(b) contain colors of equal hue but different saturation and brightness. We used a colorimeter (KONICA MINOLTA, CR-20) to adjust the target to the desired color. The following procedures were repeated until the printed color reached the desired color: setting color parameters on PC, printing out to paper, and measuring the printed color patch using the colorimeter. The target colors for discrimination in this study are dull, bright, and vivid red, as well as dull, bright, and vivid green.
The dull colors are used as reference colors. The bright color has higher brightness than the dull color, and the vivid color has higher saturation than the dull color.

RGB information obtained as the average value within the white frame was converted to the HSV color space and the DO color plane, and the converted values were evaluated. 

\section{Results}
\subsection{Discrimination of hue}
\label{sbsec:Discrimination of hue}
\begin{figure}[tb]
    \centering
		\includegraphics[keepaspectratio=true,width=1.0\linewidth]{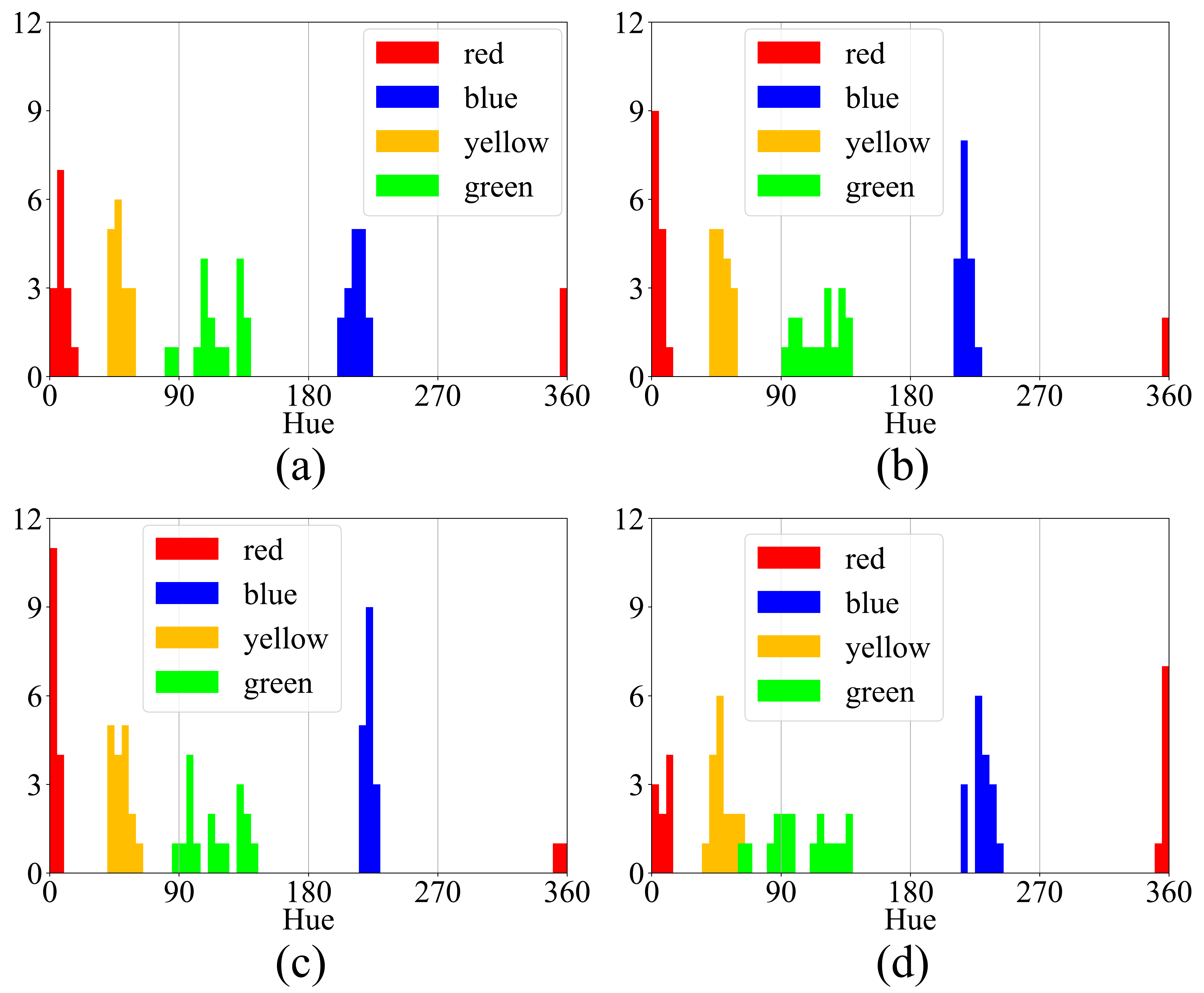}
	\caption{Histogram of hue computed by some of the models when the beverage cartons were the visual targets. (a) C/S retinex model with $\gamma=3$. (b) C/S retinex model with $\gamma=6$. (c) N-R C/S retinex model. (d) Gray-world algorithm.}
	\label{fig:paperPack_H}
\end{figure}

Fig. \ref{fig:paperPack_H} shows the histogram of hue in the HSV color space computed by some of the models in the experiment in which beverage cartons were used. The color of each bin corresponds to each color of the target carton. The sum of the histograms for each target is 17, which corresponds to 17 different colors of illumination.

\begin{figure}[bt]
    \centering
		\includegraphics[keepaspectratio=true,width=1.0\linewidth]{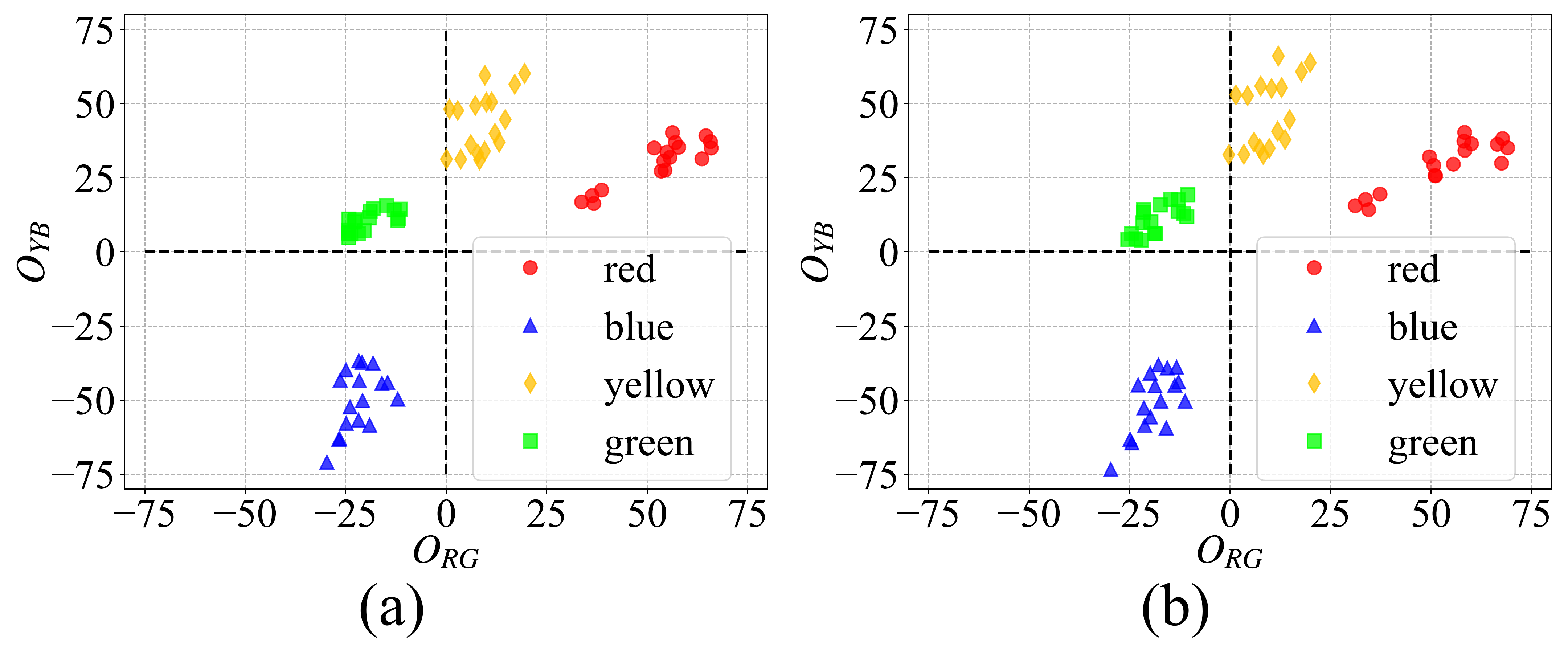}
	\caption{$O_{RG}$ and $O_{YB}$ computed by some of the models when the beverage cartons were the visual targets. (a) C/S retinex model with $\gamma=6$. (b) N-R C/S retinex model.}
	\label{fig:paperPack_RGYB}
\end{figure}

Fig. \ref{fig:paperPack_RGYB} shows $O_{RG}$ and $O_{YB}$ in the DO color plane computed by some of the models in the experiment in which beverage cartons were used. The color of each plot corresponds to each color of the target carton. Also, for each target color, 17 points are plotted, which correspond to the values obtained under 17 different illuminations.

\begin{figure}[bt]
    \centering
		\includegraphics[keepaspectratio=true,width=1.0\linewidth]{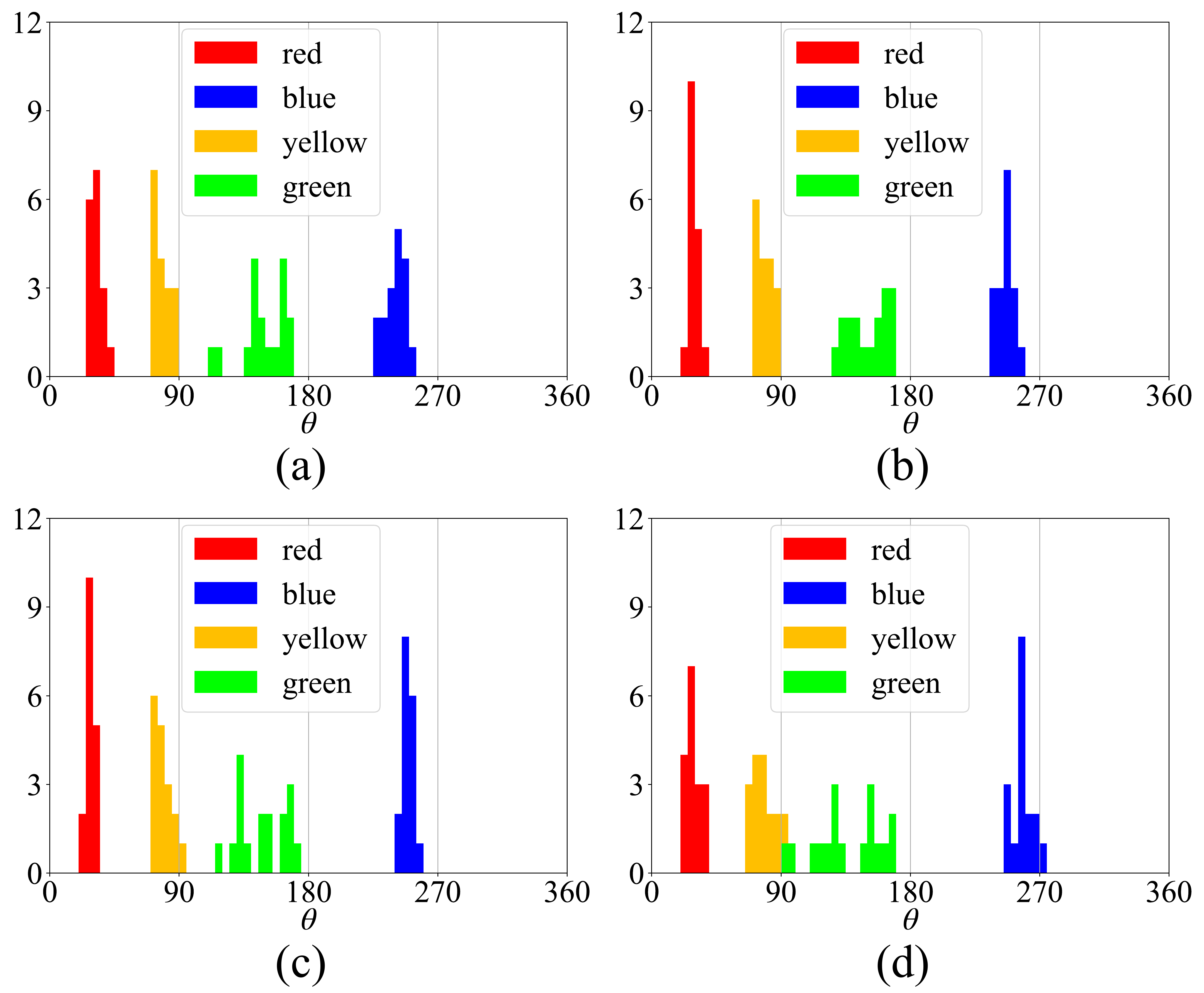}
	\caption{Histogram of $\theta$ in the DO color plane computed by some of the models when the beverage cartons were the visual targets. (a) C/S retinex model with $\gamma=3$. (b) C/S retinex model with $\gamma=6$. (c) N-R C/S retinex model. (d) Gray-world algorithm.}
	\label{fig:paperPack_RGYB_theta}
\end{figure}

Fig. \ref{fig:paperPack_RGYB_theta} shows the histogram of $\theta$ in the DO color plane computed by some of the models in the experiment in which beverage cartons were used. The color of each bin corresponds to each color of the target carton. The sum of the histograms for each target is 17, the same as in Fig. \ref{fig:paperPack_H}.

Table \ref{tab:paperPack_hueSep} shows the FC for hue discrimination. The comparison is based on the discrimination of the following four color-pairs that are close in hue: red/yellow, yellow/green, green/blue, and blue/red. In the table, Log($\gamma:0$), Log($\gamma:3$), Log($\gamma:6$), Log($\gamma:9$) represent the C/S retinex model with $\gamma=0, 3, 6, 9$, respectively. Linear represents the C/S retinex model that uses the linear function in place of the logarithmic function. GW represents the GW algorithm. The C/S retinex model with the logarithmic function showed high FC for $\gamma$ = 6. We examined the value of FC for smaller steps of $\gamma$ around 6 (not shown in the table), but the value did not change significantly and was constantly high. On the other hand, the model with linear function showed the lowest FC for the three color-pairs. The N-R C/S retinex model showed maximum FC for red/yellow and blue/yellow discrimination and maintained high values without any parameter tuning. The results indicate that the choice of light intensity encoding function has strong influence on the color discrimination performance of the C/S retinex model. The performance of the GW algorithm was low for all the color-pairs.
\begin{table}[t] 
	\caption{Fisher criteria for hue discrimination. Numbers in bold denote maximum values. Underlined numbers denote minimum values.}
    \centering
		\begin{tabular}{l|l|cccc} \hline
			\multicolumn{2}{c|}{ } & red \& & yellow & green & blue \\
			\multicolumn{2}{c|}{ } & yellow & \& green & \& blue & \& red \\ \hline
			\multirow{6}{*}{C/S}	& Log $(\gamma:0)$	& 15.96 & 12.84 & 27.17 & 210.23 \\
				& Log $(\gamma:3)$	& 33.11 & 15.55 & 31.26 & 354.84 \\
				& Log $(\gamma:6)$	& 49.84 & \textbf{20.15} & \textbf{45.44} & 657.84 \\
				& Log $(\gamma:9)$	& 7.32 & 9.07 & 16.86 & 72.11 \\
				& Linear			& \underline{3.57} & 6.65 & \underline{10.08} & \underline{33.57} \\
				& N-R				& \textbf{52.24} & 12.62 & 34.63 & \textbf{747.74} \\ \cline{1-2}
			\multicolumn{2}{c|}{GW} & 27.18 & \underline{5.49} & 29.40 & 257.67 \\ \hline
		\end{tabular}
	\label{tab:paperPack_hueSep}
\end{table}
%

Table \ref{tab:paperPack_thetaSep} shows the FC for $\theta$ discrimination. $\theta$ in the DO color plane corresponds to hue in the HSV color space. The comparison is based on the discrimination of the same color-pairs as Table \ref{tab:paperPack_hueSep}. The overall trend was the same as in Table \ref{tab:paperPack_hueSep}. The FC of models except for the N-R C/S retinex model were in some cases larger and in some cases smaller than the FC obtained using the HSV color space. On the other hand, the FC of the N-R retinex model improved for all color pairs and reached a maximum value for two color pairs.
\begin{table}[!ht] 
	\caption{Fisher criteria for $\theta$ discrimination. Numbers in bold denote maximum values. Underlined numbers denote minimum values.}
    \centering
		\begin{tabular}{l|l|cccc} \hline
			\multicolumn{2}{c|}{ } 	& red \& & yellow & green & blue \\
			\multicolumn{2}{c|}{ } 	& yellow & \& green & \& blue & \& red \\ \hline
			\multirow{6}{*}{C/S}	& Log $(\gamma:0)$	& 21.56 & 15.29 & 25.06 & 221.57 \\
				& Log $(\gamma:3)$	& 40.86 & 18.54 & 28.81 & 322.60 \\
				& Log $(\gamma:6)$	& 54.97 & \textbf{25.62} & \textbf{47.40} & 597.93 \\
				& Log $(\gamma:9)$	& 9.31 & 10.24 & 16.57 & 81.67 \\
				& Linear			& \underline{4.41} & 7.36 & \underline{9.84} & \underline{37.23} \\
				& N-R				& \textbf{58.83} & 15.70 & 36.17 & \textbf{774.81} \\ \cline{1-2}
			\multicolumn{2}{c|}{GW}	& 30.98 & \underline{6.03} & 27.04 & 250.87 \\ \hline
		\end{tabular}
	\label{tab:paperPack_thetaSep}
\end{table}

\subsection{Discrimination of saturation and brightness}
\label{sbsec:Discrimination of saturation and brightness discrimination}
\begin{figure}[bt]
    \centering
		\includegraphics[keepaspectratio=true,width=1.0\linewidth]{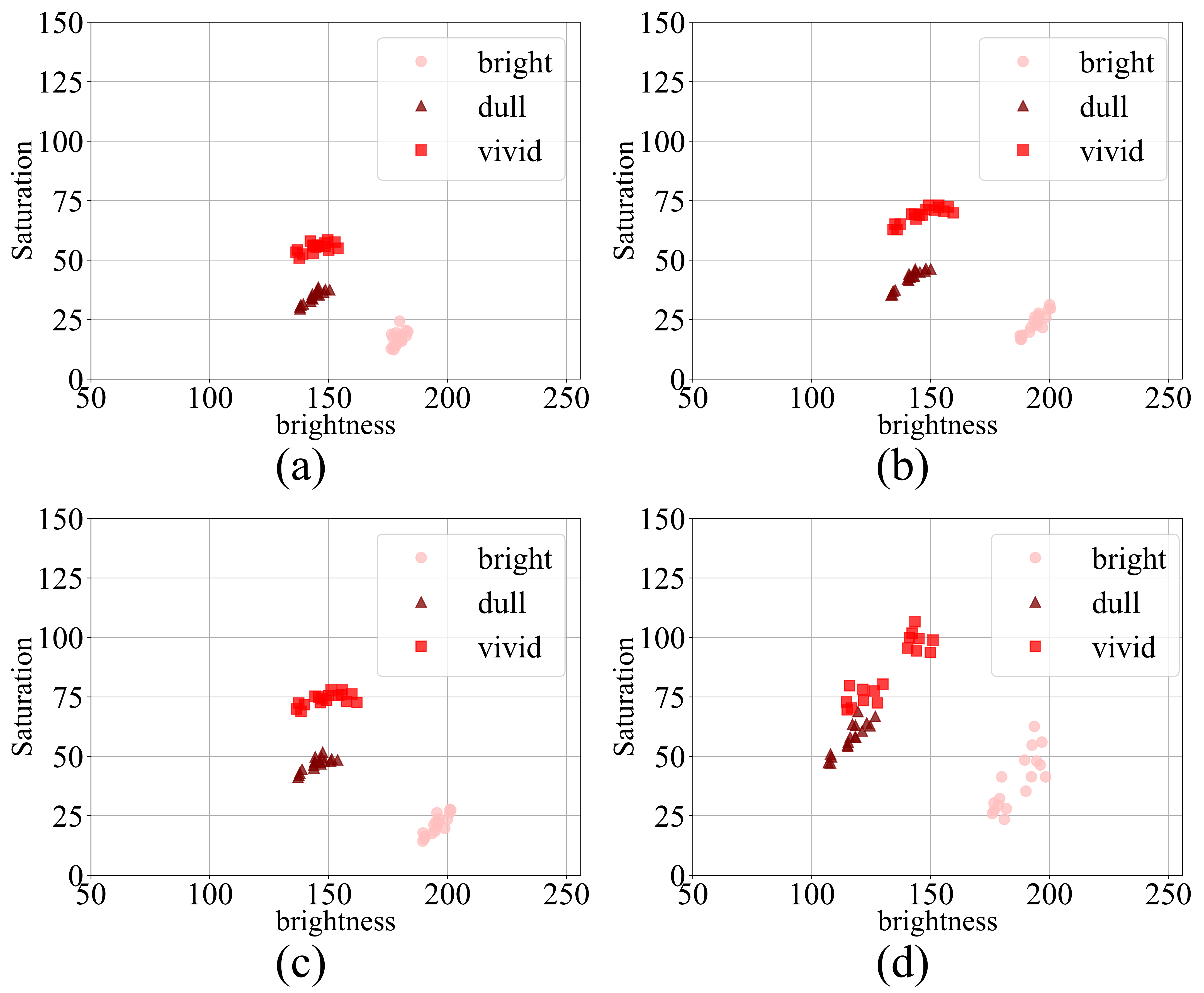}
	\caption{Brightness and saturation computed by some of the models when the red color patches were visual targets. (a) C/S retinex model with $\gamma=3$. (b) C/S retinex model with $\gamma=6$. (c) N-R C/S retinex model. (d) Gray-world algorithm.}
	\label{fig:redTarget_SV}
\end{figure}

Fig. \ref{fig:redTarget_SV} shows the brightness and saturation computed by some of the models for each experiment in which the red color patches were the visual targets. The color of each plot corresponds to the color of each target. For each color, 17 points are plotted, the same as in the experiments of hue.

\begin{figure}[bt]
    \centering
		\includegraphics[keepaspectratio=true,width=1.0\linewidth]{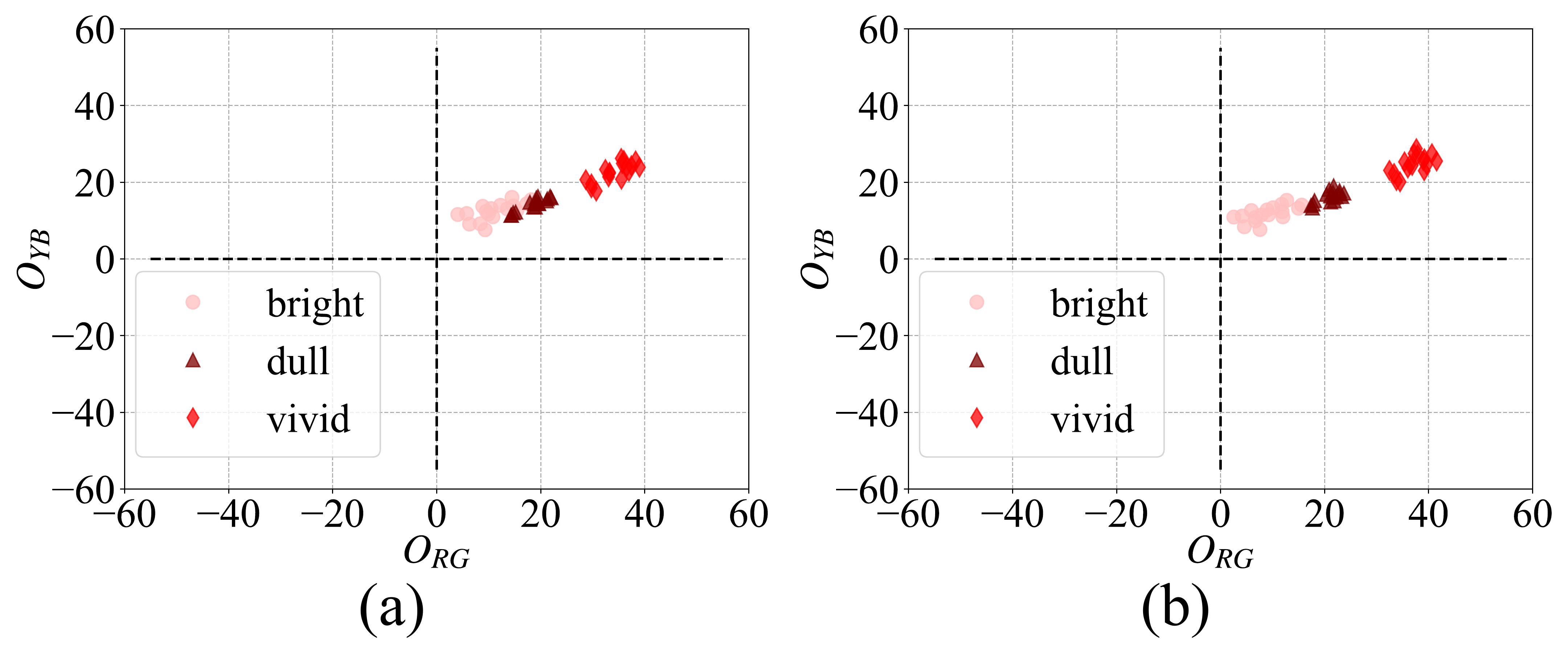}
	\caption{$O_{RG}$ and $O_{YB}$ computed by some of the models when the red color patches were the visual targets. (a) C/S retinex model with $\gamma=6$. (b) N-R C/S retinex model.}
	\label{fig:redTarget_RGYB}
\end{figure}

Fig. \ref{fig:redTarget_RGYB} shows $O_{RG}$ and $O_{YB}$ in the DO color plane computed by some of the models for each experiment in which the red color patches were the visual targets.
The color of each plot corresponds to the color of each target color patch. For each shape, 17 points are plotted, the same as in Fig. \ref{fig:redTarget_SV}.

\begin{figure}[bt]
    \centering
		\includegraphics[keepaspectratio=true,width=1.0\linewidth]{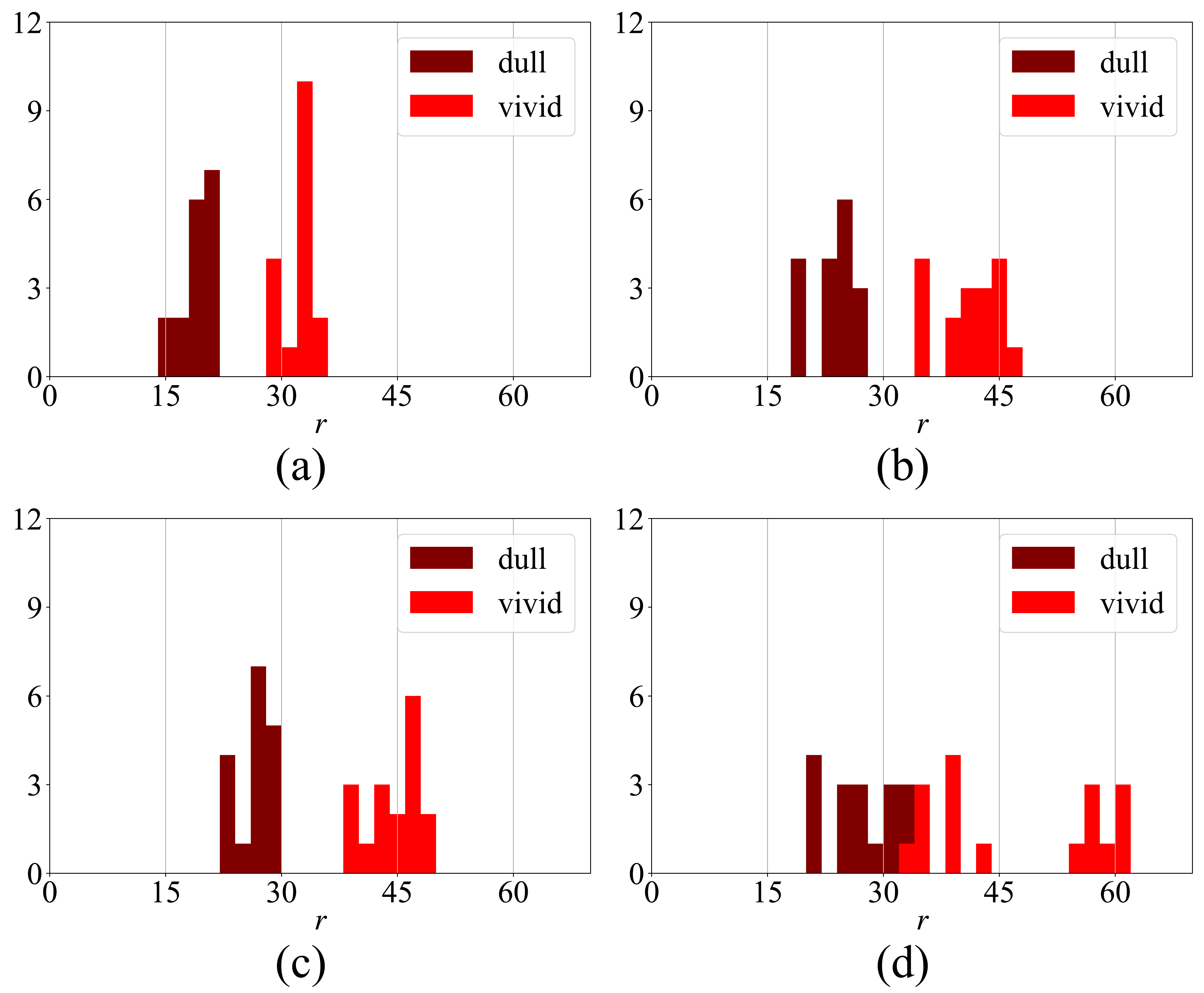}
	\caption{Histogram of $r$ in the DO color plane computed by some of the models when the red color patches were the visual targets. (a) C/S retinex model with $\gamma=3$. (b) C/S retinex model with $\gamma=6$. (c) N-R C/S retinex model. (d) Gray-world algorithm.}
	\label{fig:redTarget_RGYB_r}
\end{figure}

Fig. \ref{fig:redTarget_RGYB_r} shows the histogram of $r$ in the DO color plane computed by some of the models in the experiment in which the red color patches were the visual targets. The color of each bin corresponds to each color of the target carton. The sum of the histograms for each target is 17, the same as in the experiments of hue.

Table \ref{tab:redTarget_rSVSep} shows the FC for $r$, saturation, and brightness discrimination when the red color patches were the visual targets. $r$ in the DO color plane corresponds to saturation in the HSV color space. The ability to discriminate saturation was evaluated by the FC on the discrimination between dull red and vivid red. The C/S retinex model using logarithmic function showed a larger FC for a relatively small value of $\gamma$. The results showed that the use of the HSV color space offers a better performance in red saturation discrimination for all methods compared here. The N-R C/S retinex model with the HSV color space achieved the maximum value. The ability to discriminate brightness was evaluated by the FC on the discrimination between dull red and bright red. The C/S retinex model using the logarithmic function with $\gamma=0$ showed the largest value, but the difference between models was not as large as that of saturation.
\begin{table}[bt] 
	\caption{Fisher criteria for $r$, saturation, and brightness discrimination (red patch). Numbers in bold denote maximum values. Underlined numbers denote minimum values.}
    \centering
		\begin{tabular}{l|l|ccc} \hline
			\multicolumn{2}{c|}{ } 	& $r$ & saturation & brightness\\
			\multicolumn{2}{c|}{ } 	& vivid \& dull & vivid \& dull & bright \& dull \\ \hline
			\multirow{6}{*}{C/S}	& Log $(\gamma:0)$	& 23.12 & 25.63 & \textbf{74.25} \\
				& Log $(\gamma:3)$	& \textbf{23.80} & 37.52 & 72.76 \\
				& Log $(\gamma:6)$	& 14.13 & 29.85 & 67.40 \\
				& Log $(\gamma:9)$	& 10.38 & 20.86 & 60.33 \\
				& Linear			& 9.84 & 19.26 & 58.28 \\
				& N-R				& 22.15 & \textbf{54.30} & 67.55 \\ \cline{1-2}
			\multicolumn{2}{c|}{GW}	& \underline{3.09} & \underline{4.06} & \underline{51.27} \\ \hline
		\end{tabular}
	\label{tab:redTarget_rSVSep}
\end{table}
\begin{table}[bt] 
	\caption{Fisher criteria for $r$, saturation, and brightness discrimination (green patch). Numbers in bold denote maximum values. Underlined numbers denote minimum values.}
    \centering
		\begin{tabular}{l|l|ccc} \hline
			\multicolumn{2}{c|}{ } 	& $r$ & saturation & brightness \\
			\multicolumn{2}{c|}{ } 	& vivid \& dull & vivid \& dull & bright \& dull \\ \hline
			\multirow{6}{*}{C/S}	& Log $(\gamma:0)$	& \textbf{20.75} & 12.05 & 148.51 \\
				& Log $(\gamma:3)$	& 18.76 & \textbf{12.75} & 145.54 \\
				& Log $(\gamma:6)$	& 11.62 & 8.79 & 139.90 \\
				& Log $(\gamma:9)$	& 7.88 & 6.71 & 135.57 \\
				& Linear			& 6.51 & 6.05 & 139.85 \\
				& N-R				& 11.86 & 8.48 & \textbf{167.01} \\ \cline{1-2}
			\multicolumn{2}{c|}{GW}	& \underline{4.76} & \underline{4.24} & \underline{123.76} \\ \hline
		\end{tabular}
	\label{tab:greenTarget_rSVSep}
\end{table}
Table \ref{tab:greenTarget_rSVSep} shows the FC for $r$, saturation, brightness discrimination when the green color patches were the visual targets. The color pairs to be compared are the same in Table \ref{tab:redTarget_rSVSep}, which is the results of the experiment using the red patches. The FC for $r$ and saturation discrimination showed a larger value when the C/S retinex model using logarithmic function with a relatively small value of $\gamma$. In contrast to the experiment with the red patches, where the use of the HSV color space gave higher FC, the use of the DO color plane gave higher FC for all methods compared here.
\section{Conclusion}
\label{sec:Conclusion}
In this study, we evaluated the color discrimination performance of the C/S retinex model in terms of robot vision applicability, as well as effects of the light intensity encoding function and color space on the color discrimination performance. 

The GW algorithm was used as a representative of conventional low-computational-cost CC algorithms. The evaluation results using FC showed that the C/S retinex model outperforms the GW algorithm in discriminating all color attributes as long as the light intensity encoding function is set appropriately. 

The parameter $\gamma$ of the logarithmic function that maximizes the FC depended on the hue and other factors of the target. Adaptive adjustment of $\gamma$ is ideal, but another study focusing on this adjustment is needed. Without the need for adaptive adjustment, the N-R function consistently exhibited a high FC. Since the N-R function has an adaptive parameter $I_{h}$, adaptive control of this parameter could further improve discrimination performance. 

Differences in the color space used also affected discrimination performance. For the C/S retinex using the logarithmic function, the color space in which the FC was greater varied depending on the attribute of the color to be discriminated. For the C/S retinex using the N-R function, the use of the DO color plane resulted in a larger FC, except for red saturation discrimination. 

In summary, the combination of the N-R function, the DO color plane, and the retinex model is one of the most suitable CC algorithms at least in the environment used in this study.

The environment used in this study simulated the varied illumination colors of a real environment, and we believe that the illumination color variations achieved by color combinations using two color-variable LEDs cover a wide range of challenging illumination colors that robot vision may face in a real applications. However, it is important to note that the color represented by the C/S retinex model depends on the average color in the area of the “surround” Gaussian filter, which has the larger SD. If the average color is constant within the area of the Gaussian filter, high CC performance as shown in this study can be expected even in real environments. On the other hand, if the background color of the target does not match this condition, the represented color of the target would change to some degree. Therefore, it will be necessary to change the SD of the "surround" Gaussian filter and the assumed variation range of the target color represented by the model, depending on the scenario.


\bibliography{ref_yamada.bib} 
\bibliographystyle{IEEEtran} 


\end{document}